\pdfoutput=1

\documentclass[11pt]{article}

\usepackage{acl}
\usepackage{enumitem}
\usepackage{breqn}
\usepackage{devanagari}
\usepackage{amsmath}
\usepackage{times}
\usepackage{latexsym}
\usepackage{graphicx}
\usepackage[T1]{fontenc}

\usepackage[utf8]{inputenc}

\usepackage{microtype}

\title{Elevating Code-mixed Text Handling through Auditory Information of Words}

\author{Mamta, Zishan Ahmad \and Asif Ekbal\\
      Department of Computer Science 
 and Engineering \\
 Indian Institute of Technology Patna, India \\
   \texttt{\{mamta\_1921cs11, 1821cs18, asif\}@iitp.ac.in}
  }
  
\begin{document}
\maketitle
\begin{abstract}
With the growing popularity of code-mixed data, there is an increasing need for better handling of this type of data, which poses a number of challenges, such as dealing with spelling variations, multiple languages, different scripts, and a lack of resources. Current language models face difficulty in effectively handling code-mixed data as they primarily focus on the semantic representation of words and ignore the auditory phonetic features. This leads to difficulties in handling spelling variations in code-mixed text. In this paper, we propose an effective approach for creating language models for handling code-mixed textual data using auditory information of words from SOUNDEX. Our approach includes a pre-training step based on masked-language-modelling, which includes SOUNDEX representations (SAMLM) and a new method of providing input data to the pre-trained model. Through experimentation on various code-mixed datasets (of different languages) for sentiment, offensive and aggression classification tasks, we establish that our novel language modeling approach (SAMLM) results in improved robustness towards adversarial attacks on code-mixed classification tasks. Additionally, our SAMLM based approach also results in better classification results over the popular baselines for code-mixed tasks. We use the explainability technique, SHAP (SHapley Additive exPlanations) to explain how the auditory features incorporated through SAMLM assist the model to handle the code-mixed text effectively and increase robustness against adversarial attacks \footnote{Source code has been made available on \url{https://github.com/20118/DefenseWithPhonetics}, \url{https://www.iitp.ac.in/~ai-nlp-ml/resources.html\#Phonetics}}.
\end{abstract}

\section{Introduction}
The proliferation of code-mixed content on social media platforms among multilingual communities around the globe has been widely observed in recent years. It has been established that handling code-mixed content for information retrieval or classification poses a unique set of challenges. These challenges become even more prominent when a language is written in a different script during code-mixing. Since there are no formal spelling standards for a word in a different script, there can be large variations in spellings (Eg: `{\dn hA\1}' (yes) in Hindi can be written as `haan’, `haa’, `ha’ etc.). These spelling variations depend on many socio-cultural factors, such as dialect, accent, and region \cite{crystal1987cambridge}. It has been noted that a significant portion of the code-mixed content present on social media platforms is Romanized, which presents a challenge in terms of processing and analysis due to the lack of following a standardized Romanization method. This lack of standard leads to many complexities and is one of the major roadblocks in training a reliable and robust code-mixed NLP system \cite{chittaranjan2014word,vyas2014pos}. Managing such variations within text data are typically achieved through pre-processing techniques, such as data augmentation and normalization \cite{kusampudi2021sentiment}, which necessitates the utilization of human-annotated dictionaries and can entail a significant investment in manual annotation efforts. It has been observed that traditional techniques for processing and analysis of code-mixed content may prove ineffective in cases where the spelling of a word varies from those present in the corpus or dictionary \cite{das2022advcodemix}.

Although transformer based pre-trained models \cite{devlin2018bert,liu2019roberta} have proven to be largely effective for most of the tasks in Natural Language Processing (NLP) \cite{ mamtaetal-2022-hindimd,sun2019utilizing}, it has been shown that even such models are not robust enough to handle small perturbations in spelling \cite{das2022advcodemix}.
Such perturbations have been used to perform adversarial attacks on even the transformers based language models. Adversarial attack entails making small human imperceptible perturbations to the input to mislead the models. The first study in this direction proposed three adversarial attacks based on phonetic perturbations to test the limits of a code-mixed text classifier. 
 In this study, it was found that the BERT \cite{devlin2018bert} model was vulnerable to such phonetic perturbations.
\citet{van1987rows} found that phonetically similar spelling variations of a word are often imperceptible to humans. 
For example, words \textit{acha} (meaning `okay'), \textit{acchha}, and \textit{achha} have similar sounds when spoken. These properties of words are known as SMS property (similar sound, similar meaning, different spellings) \cite{le2022perturbations}. 

In this paper, we focus on incorporating the auditory phonetic (AP) features of words along with their semantic features 
in language models. We hypothesize that a model trained by utilizing these features would be agnostic to subtle spelling variations. These variations are often found to be the Achilles heel of deep learning systems and such variations are exploited during adversarial attacks. Incorporating these features would also lead to building better and more robust classifiers for code-mixed input.

To obtain the AP features, we utilize the SOUNDEX algorithm \cite{stephenson1980methodology}. This algorithm encodes the SMS property of words. In this encoding, the words acha (ok), achha, and acchha have the same encoding vectors (A200). To embed these phonetic properties, we propose two novel language modeling approaches named SOUNDEX Language Modelling (SMLM) and SOUNDEX Aligned Masked Language Modeling (SAMLM) that are able to map between the semantic and auditory properties of words in a text.
We use these approaches to pre-train BERT and RoBERTa models. 
We then fine-tune our pre-trained models on downstream classification tasks based on code-mixed Hinglish (Hindi+English) and Benglish (Bengali+English) datasets. We perform phonetic perturbation-based attacks following \citet{das2022advcodemix} and find that our SMLM and SAMLM pre-trained models are more robust to such adversarial attacks. We observe a lower drop in performance in both models when compared to the base BERT and RoBERTa models after the attack. Additionally, we also observe a improvement in classification scores on the downstream tasks of code-mixed text classification in both languages. 

However, these models lack transparency, which makes it difficult to understand their actual decision process. Hence, we exploit the model explainability to analyze the decision process of our models by extracting the terms responsible for the final prediction. For this purpose, the explainability technique, SHAP (SHapley Additive exPlanations) \cite{lundberg2017unified} is used. To the best of our knowledge, this is the very first attempt towards utilizing AP properties to enhance the robustness of models while dealing with code-mixed datasets. The key contributions of this work are as follows:
 \begin{itemize}[noitemsep]
     \item To align the semantic and AP features of a text, we propose two novel pre-training steps for BERT and RoBERTa, \textit{viz.} (i). SOUNDEX Language Modeling (SMLM), and (ii). SOUNDEX Aligned Language Modeling (SAMLM).  
     \item We illustrate the effectiveness of the proposed technique as a defense against adversarial attacks without the need for re-training the model on adversarial attack samples.
     \item Extensive experiments on Hinglish and Benglish code-mixed datasets show that using our pre-training steps (SMLM and SAMLM) results in better classification performance for code-mixed settings. 
      \item To better understand how the utilization of AP features affects the decision process of BERT and RoBERTa, we harness the model explainability technique, SHAP.
 \end{itemize}

\section{Related Work}
Transformers-based pre-trained models have achieved remarkable success in a wide range of NLP tasks \cite{li2019entity,raffel2020exploring,mamtaecir23}. However, several studies have shed light on vulnerabilities of these models \cite{sun2020adv}. \citet{jin2020bert} propose a black-box algorithm to attack BERT model with the help of closet synonyms. But it can lead to unnatural sentences because the synonym may not fit the context of the sentence. To overcome this limitation, authors in \cite{garg2020bae,li2020bert,mondal2021bbaeg,m-ekbal-2022-adversarial} proposed to use a masked language model (BERT or RoBERTa) for replacements or insertions. There are numerous studies to enhance adversarial robustness using data augmentation, adversarial training \cite{morris2020textattack}, etc. Data augmentation requires manual human efforts and adversarial training requires re-training models on adversarial data which is costly. However, all these attempts are for the high-resource English language except \citet{m-ekbal-2022-adversarial}. 

Increasing phenomena of code-mixing on social media platforms have also motivated researchers to analyze the adversarial robustness of code-mixed models. Authors in \cite{das2022advcodemix} exposed the vulnerability of code-mixed classifiers by performing an adversarial attack based on subword perturbations, character repetition, and word language change. However, there is no attempt to enhance the adversarial robustness of code-mixed text classifiers against these perturbations. This motivated us to develop a robust model to handle adversarial perturbations for code-mixed text.

Researchers analyzed the behaviour of pre-trained language models (PMLM) for different languages and attempted to enhance their performance on the downstream tasks. For example, \citet{hande2021benchmarking} conducted experiments on Tamil, Kannada, and Malayalam scripts and observed that multilingual models perform better than monolingual models. \citet{mamta2023transformer} proposed a multilingual framework to fine-tune BERT in shared private fashion to transfer knowledge between code-mixed and English languages. \citet{rathnayake2022adapter} performs adapter-based fine-tuning of PMLMs for code-mixed text classification. However, their focus is not on handling phonetic perturbations based adversarial attacks. 
There are a few attempts to enrich the representation of pre-trained models like BERT in the speech domain. For example, \citet{sundararaman2021phoneme} proposed a BERT-style language model, referred to as PhonemeBERT that learns a joint language model with phoneme sequence and Audio Speech Recognition (ASR) errors to learn phonetic-aware representations that are robust to ASR errors. They introduced noise to speech (noise related to door opening, aircaraft, etc.) and handled them using phoneme sequences. However, our task is different from above in the following aspects: (i). our focus is to enhance adversarial robustness of code-mixed classifiers against adversarial attacks; (ii). our proposed approach is tuned to handle textual perturbations in code-mixed data rather than perturbations in speech signals.

\section{Threat Model}
\label{sec:threat}
Our target models are BERT and RoBERTa based code-mixed text classifiers due to their huge success in many NLP tasks \cite{liu2019multi,xu2019bert}. An adversary attempts to mislead the target models by generating adversarial samples to make wrong classification decision. 
\\
\textbf{Adversary's goal:}
Given an input sentence $S$, consisting of $n$ tokens ${w_1, w_2, w_3,\ldots, w_n}$, with ground truth label $y$, and a target model $M(S) = y$, the goal of the adversary is to perform an un-targeted attack, i.e., find an adversarial sample $S_{adv}$, causing $M$ to perform misclassification, i.e., $M(S) != y$. Adversaries attack the model using phonetic perturbations in line with the prior work \citet{das2022advcodemix}.
\\
\textbf{Design goals:}  Based on the aforementioned adversary model, our proposed framework (SMLM and SAMLM) must meet the 
robustness and accuracy requirements.
\begin{itemize}[noitemsep]
    \item \textit{Robustness:} SMLM and SAMLM should be robust to adversarial perturbations. They should correctly classify the adversarial samples generated by the adversary.
    \item \textit{Accuracy:} SMLM and SAMLM should handle the spelling variations in real code-mixed datasets. As a result, accuracy on actual code-mixed test sets should increase.
\end{itemize}

\section{Methodology}
\label{sec:method}
\begin{figure*}[h]
  \centering
  \includegraphics[scale=0.51]{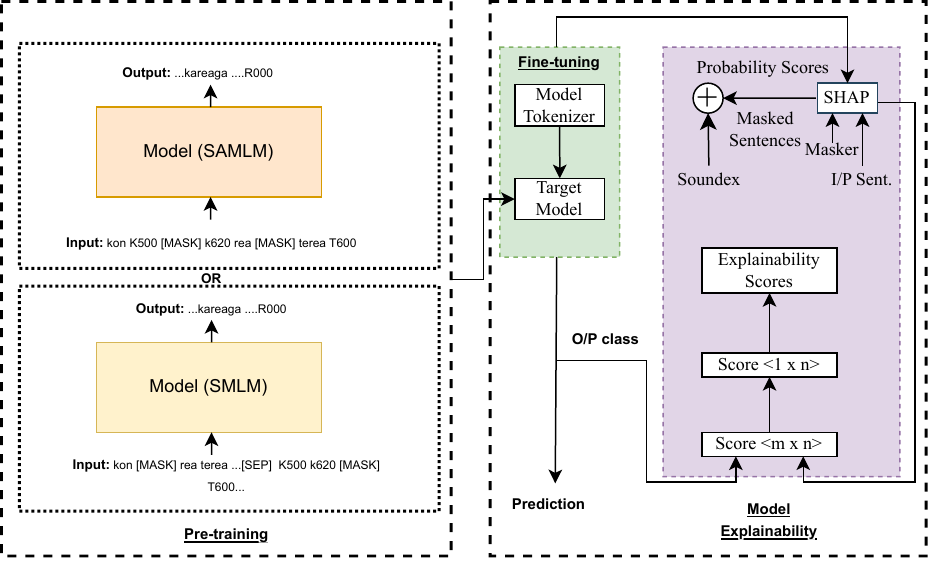}
  \caption{Proposed process flow of our proposed methodology. The pre-training of BERT or RoBERTa is done using SMLM or SAMLM, which is fine-tuned on the downstream classification task.}
  \label{archii}
  
\end{figure*}

Our objective is to equip the pre-trained models to increase their robustness against adversarial attacks and handle phonetic spelling variations in code-mixed datasets. 
The detailed flow of our proposed approach is shown in Figure \ref{archii}. There are 3 main components, \textit{viz.} pre-training, fine-tuning, and model explainability. First, we pre-train the models (BERT and RoBERTa) to incorporate auditory features, followed by task-specific fine-tuning. 
Finally, the model explainability component explains the decision process of our proposed approach and illustrate the effectiveness of our proposed approach. It analyzes how the adversarial attacks and phonetic spelling variations are handled by our proposed models qualitatively. 

\textbf{SOUNDEX Algorithm}
\label{subsec:sound}
To encode the sound of a word, we utilize the case-insensitive SOUNDEX algorithm \cite{stephenson1980methodology}. It indexes word based on their sound rather than their spelling. To assign sound encoding to a given word, SOUNDEX first retains the initial character followed by the removal of all vowels. It then maps the remaining characters one by one to a digit with the help of predefined rules.
In this manner, SOUNDEX assigns the same encoding (A200) to different variations \textit{acha} (good), \textit{acchha}, and \textit{acchha}. However, in code-mixed language, same word might have different meanings in two or more languages. For example, Hindi word \textit{yar} (friend) and year will share the same SOUNDEX vector (Y600). But these words have different meanings. Our proposed approach takes care of this limitation of SOUNDEX.

\subsection{Pre-training}
\textbf{SOUNDEX Masked Language Sound Modelling (SMLM)} 
\label{subsec:smlm}
A common denominator between spelling variations of a word is the similar AP property of the variations. Modeling this auditory property to the model would increase the model's robustness and help in better classification of code-mixed text. To incorporate this property, we use SOUNDEX encoding in our language model along with the usual contextual word encoding. The SOUNDEX sequence $A=\{s_1,s_2,...,s_n\}$ for the sentence $S=\{t_1,t_2,...,t_n\}$ ($t_i$ is the WordPiece token obtained by passing the sentence to the model tokenizer) is obtained and a joint input sequence $IP = [t_1, t_2, ..., t_n, [SEP], s_1, s_2,...s_n]$ is formed. We follow the masked-language-modeling approach proposed by \citet{devlin2018bert} on the sequence. In order to train a deep bidirectional representation, we simply mask some percentage of the input tokens at random and then predict those masked tokens at the output layer of the model. The masked tokens can be either from the subsequence $S$ or $A$. When a token from $S$ is masked it predicts the word attending to both the contextual information in $S$ and the auditory information in $A$. In this manner, the model would learn to predict the semantically correct word and auditory correct word. When a token from the subsequence $A$ is masked, the model would learn to predict the auditory SOUNDEX encoding of the respective word in the input sentence. In this way, SMLM can handle the limitation of SOUNDEX. 

In all of our experiments, we mask 15\% WordPiece tokens in each sequence at random. The final loss $L$ at the output layer is given in Equation \ref{eq:smlmloss}.

\begin{equation}
\label{eq:smlmloss}
\resizebox{\hsize}{!}{%
   $ L = -\frac{1}{N} \sum_{i=1}^{N} \log p(x_i| x_{1}, x_{2}, .., x_{i-1}, x_{i+1},..,x_{N})$}
\end{equation}

Here, $N$ is the total number of masked tokens in the input sequence ($IP$ in our case), $x_i$ is the masked i-th token, and $p(x_i| x_{1}, x_{2}, ..., x_{i-1},$ $ x_{i+1},...,x_{N})$ is the probability of the i-th token, conditioned on all the other tokens in the sequence.

\paragraph{SOUNDEX Aligned Masked Language Modelling}
\label{subsec:samlm}
Although SMLM incorporates auditory properties along with the semantic characteristics of the word, both of these properties might not always align. This is because the text $S$ and auditory $A$ sequences are appended one after the other. In each of the sequences, $t_i$ and $s_i$ can be split into multiple tokens (during WordPiece tokenization), making this alignment even more difficult. For better alignment between word and SOUNDEX tokens, we propose a SOUNDEX Aligned Masked Language Modelling (SAMLM). In this method, instead of appending the sequence one after the other, we make a new input sequence by interleaving the two sequences $IP_1 = \{t_1, s_1, t_2, s_2,...,t_n, s_n \}$. This input sequence takes care of the alignment of auditory tokens with the word tokens, which would ensure more robustness in the model in case of adversarial attacks and natural spelling variations in the code-mixed text. In addition, SAMLM's semantic alignment can take care of the limitation of SOUNDEX more effectively.

\subsection{Fine-tuning}
\label{subsec:finetune}
Once the model (BERT or RoBERTa) is pre-trained using our proposed approaches (SMLM and SAMLM), the model is fine-tuned for the downstream classification tasks. For models trained with the SMLM approach, we created the input sequence $IP_{smlm} = \{[CLS], t_1,t_2,...,t_n, [SEP], s_1,s_2,...,s_n \}$. Similarly for models trained with the SAMLM approach, the prepared input sequence is $IP_{samlm} = \{[CLS], t_1,$ $s_1, t_2,s_2...,t_n,s_n \}$. This input sequence is passed to the model and from the pre-final layer of the model $[CLS]$ representation is fed into an output layer for the classification tasks. 

\subsection{Model Explainability}
Model explainability component is introduced to understand how auditory features help the model in improving its robustness and accuracy. We use Shapely algorithm to determine the relevance of each word in a given sentence, against the target model (BERT and RoBERTa). It calculates the relevance score (known as Shapley value) for each word based on possible coalitions of words for a particular prediction \cite{lundberg2017unified} \footnote{More details can be found in \cite{lundberg2017unified}}. 

We create an explicit word masker which tokenizes the sentence into fragments containing words, and which is then used to mask words in SHAP (here mask refers to hiding a particular word from the sentence). The input sentence along with the designed masker is passed to SHAP which generates various masked combinations of the sentence. 
These masked sentence fragments are further passed to the model tokenizer. 
We further concatenate the SOUNDEX encoding to the masked combinations 
for better prediction scores as shown in Figure \ref{archii}. This concatenation further helps Shapley to compute the relevance scores of words based on semantic and auditory features. Both the model tokenizers (BERT and RoBERTa) convert the words to subwords and generate input, segment, and mask embeddings for each subword unit, and generate the final representation by performing a summation of all three embeddings \cite{devlin2018bert}. 
Finally, this combined representation of these vectors for each masked version is passed to the target model to obtain the output probabilities, which are further returned to SHAP to obtain the relevance of each word for the final prediction. 

\section{Experimental Setup and Results}
We use BERT-base and RoBERTa-base as target models for each task. To access the experimental evaluation of our proposed approach, we conduct extensive experiments on code-mixed Hinglish and Benglish language datasets. For Hinglish, we conduct experiments on two benchmark datasets related to offensive \cite{mathur2018did} and sentiment analysis \cite{joshi2016towards}. For Benglish, we conduct experiments on aggression analysis data \cite{bhattacharya2020developing}. Similarly, for pre-training task, we use a total of 33,014 Hinglish sentences and 6,149 Benglish sentences. \footnote{More details are provided in the Appendix \ref{sec:appendix:impdata}.}

\subsection{Baselines}

\textbf{Vanilla classifiers (VC):} We fine-tune the vanilla BERT and RoBERTa (henceforth referred to as VCBERT and VCRoBERTa) pre-trained models on the downstream tasks with only the word sequences as input.

\textbf{Vanilla Masked Language Modelling pre-trained classifiers (VMLM):} We pre-train BERT and RoBERTa on real code-mixed Hinglish/Benglish datasets (We henceforth refer to these pre-trained models as VMLMBERT and VMLMRoBERTa). After pre-training, the model is fine-tuned on their respective language datasets for the downstream tasks. During fine-tuning, only word sequences are considered as input.

 \textbf{PhoneMLM classifiers:} We pre-train BERT and RoBERTa on word and phoneme sequences. Phoneme sequences are appended at the end of word sequence separated by `[SEP]' token following \citet{sundararaman2021phoneme}. Next, each model is fine-tuned on the downstream task using both word and phoneme sequences as input. Phoneme sequences are generated using Phonemizer tool \footnote{https://pypi.org/project/phonemizer/}. 
 
 \textbf{SMLM Classifiers:} BERT and RoBERTa models are pre-trained on words and the corresponding SOUNDEX vectors. Each model is then fine-tuned on the downstream classification task.
 
 \textbf{SAMLM Classifiers:} BERT and RoBERTa models are pre-trained on words and the corresponding SOUNDEX vectors using the SAMLM strategy. Each model is then fine-tuned on the downstream classification task.

\subsection{Experimental Results}
\label{sec:exp}
We define the following two setups for the evaluation of our proposed approaches: (i). Robustness evaluation on adversarial test sets; (ii). Performance evaluation on the original test sets. We use accuracy and F1 scores to evaluate the performance on original test sets. For adversarial robustness evaluation, we use the following metrics:
\begin{itemize}[noitemsep]
    \item \textit{Before-attack-accuracy (BA) and after-attack-accuracy (AA):}  BA is calculated on the original test sets and AA score is calculated on the adversarial test set.
    \item \textit{Before-attack-F1 (BF1) and after-attack-F1 (AF1):}
    Before-attack-F1 score is calculated on the original test sets and the after-attack-F1 score is computed on the adversarial test set.
    \item \textit{Perturbation ratio (PR):} The ratio of words perturbed in the sentence to the total number of words in the sentence.
    \item \textit{Percentage drop in accuracy (PDA):} PDA is calculated as $\frac{BA-AA}{BA}$.
\end{itemize}

\subsubsection{\textbf{Evaluation on Adversarial Test Sets}}
\label{sec:evaladv}
We calculate the AA and AF1 which correspond to the accuracy and F1 scores calculated on the adversarial test sets.

\textbf{Generating Adversarial Attack Samples:}
To test the effectiveness of our proposed approach in improving the adversarial robustness of pre-trained models, we execute the black box attack following \citet{das2022advcodemix} on BERT and RoBERTa models. The attack is performed using sub-word perturbations. It makes use of a pre-existing dictionary of character groups (unigrams, bigrams, and trigrams) that can be replaced by phonetically similar character groups. 
To apply these perturbations, we first identify the important tokens, using the leave-one-out method, and then replace the important tokens with the corresponding other character groups from the dictionary. The aforementioned steps are repeated until the attack is successful. The Bengali words in our Benglish dataset consist of a mix of Romanized and Bengali script words. Since there is no dictionary available for Bengali (script) character groups for adversarial attacks, we could not perform attacks on the Bengali code-mixed dataset.

\textbf{Evaluation Results:} We define the following two setups: (i). generate attack samples by attacking the VCBERT and VCRoBERTa models and evaluate the performance on all the other models; (ii). attack individual models by generating different adversarial samples for each model. 
Results for setup 1 for sentiment and offensive tasks are depicted in Table \ref{tab:att_bert}. We observe that VCBERT and VCRoBERTa are less robust to the phonetic perturbation-based adversarial attack, resulting in a large drop in accuracy and F1 scores.
In contrast, VLMBERT and VLMRoBERTa are found to be more robust against adversarial attacks. This is expected since the VCBERT and VCRoBERTa are not pre-trained on code-mixed datasets.
It is also observed that PhoneMLM-based pre-training (of both BERT and RoBERTa) is more robust to these adversarial attacks compared to the original pre-training. It is interesting to note that even though PhoneMLM is better than the original pre-trained models, it is not always better than VMLM where the model is simply pre-trained on the code-mixed dataset (c.f. Table \ref{tab:att_bert} Offensive task). In contrast, both our proposed pre-training steps SMLM and SAMLM prove to be more robust than all the other baselines in all the tasks across the two code-mixed languages. 
\begin{table}[t]
\small
 \resizebox{0.49\textwidth}{!}
{
  \begin{tabular}{cclllllll}
  
   \hline
    \textbf{Attack Model}& \textbf{Task} &\textbf{Model}& \textbf{AA} & \textbf{AF1} 
    \\
   \hline
    VCBERT &Sentiment&VCBERT  &37.93 &25.09 \\
     Acc=67.87, F1=62.61&&VMLMBERT & 57.54&51.81\\
       &&PhoneMLM & 60.93 &56.59 \\             
         &&SMLM   &61.55&57.01\\
    &&SAMLM &\textbf{63.01}&\textbf{57.95}\\
    
VCBERT &Offensive&VCBERT &49.05&33.83 \\
     84.13, 76.00&&VMLMBERT & 70.23&54.27\\
       &&PhoneMLM &66.46 &51.50\\  
      
         &&SMLM  &70.02&55.85 \\
    &&SAMLM &\textbf{75.26}&\textbf{61.79}\\
VCRoBERTa &Sentiment&VCRoBERTa& 38.32&26.78\\
     64.90, 58.71&&VMLMRoBERTa  &62.96&58.64\\
       &&PhoneMLM& 63.35&59.53\\
         
         &&SMLM  &65.55&61.38\\
    &&SAMLM&\textbf{64.52}&\textbf{60.72}\\
VCRoBERTa &Offensive & VCRoBERTa &   54.50&37.59\\
     82.88, 73.21&& VMLMRoBERTa  & 58.42&40.12 \\
        &&PhoneMLM &59.31 &41.19\\  
         
         &&SMLM  & 67.63&54.21 & &\\  
    &&SAMLM &\textbf{68.81}&\textbf{55.92} \\
 \hline
\end{tabular}}
 \caption{Results of adversarial attack for setup 1}
  \label{tab:att_bert}
\end{table}
Setup 2 also demonstrates that our proposed SMLM and SAMLM are more resistant to adversarial attacks compared to all the other models as illustrated by AA and AF1. These results establish the fact that leveraging SOUNDEX encoding increases the robustness of BERT and RoBERTa models against adversarial attacks. 
We observe that gains for AA are smaller in setup 1 for our proposed approaches compared to setup 2 (except for offensive task). It is because, for setup 1, the attack is executed on VC models according to the token importance of VC models. It might be possible that in some of the cases, the focus of the VC model might be on different tokens compared to other models (in neutral instances). In this case, perturbations will not affect the output of other models to a larger extent.

\begin{table*}[t]
\small
\centering
 \resizebox{0.65\textwidth}{!}
{

  \begin{tabular}{cclllllllll}
    \hline
    \textbf{Model}& \textbf{Task} &\textbf{Model}& \textbf{BA} &\textbf{BF1} & \textbf{AA} &\textbf{AF1} &\textbf{PR}&\textbf{PDA}
    \\
    \hline
    BERT &Sentiment&VCBERT &  67.87&62.61 &37.93&25.09 &0.47 &44.11\\
     &&VMLMBERT  & 68.64&61.33&41.16&30.60&0.50&40.35\\ 
       &&PhoneMLM & 68.68&66.26 & 43.81&38.28&0.52&24.87 \\ 
            
         &&SMLM  & 69.33& 65.30&54.00&48.03&0.52&22.11\\
    &&SAMLM &\textbf{70.36}& \textbf{67.56}& \textbf{62.5} &\textbf{60.00}&\textbf{0.53}&\textbf{11.17}\\
    
 BERT &Offensive&VCBERT& 84.13&76.00&49.05&33.83 &0.53 &41.69\\
     &&VMLMBERT  & 88.30& 81.18& 53.45&36.79&0.55&39.46\\
       &&PhoneMLM & 87.02&79.03 &53.13 &37.70&0.55&38.94\\  
     
         &&SMLM  & 88.31&82.20&\textbf{68.83} &56.00 &0.54&\textbf{22.62}\\
    &&SAMLM &\textbf{88.93}&\textbf{83.31}&66.32&\textbf{57.05}&\textbf{0.55}&25.42\\

RoBERTa &Sentiment&VCRoBERTa & 64.90&58.71& 38.32&26.78&0.45&40.96\\
     &&VMLMRoBERTa  & 66.97& 61.93&40.25&31.07&0.49&39.89\\
       &&PhoneMLM& 66.36&61.57 &42.52&34.50&0.50&35.92\\
      
         &&SMLM  & \textbf{68.29}&\textbf{63.89}&50.77&43.41&\textbf{0.54}&25.66\\
    &&SAMLM &66.01&61.46 &\textbf{51.35}&\textbf{44.23}&0.49&\textbf{22.21}\\

RoBERTa &Offensive&VCRoBERTa &  82.88& 73.21 & 54.50&37.59&0.55&34.24\\
     &&VMLMRoBERTa  & 84.15&77.35&57.42 &39.63&0.57&31.76 \\
        &&PhoneMLM & 84.30&77.70 & 56.44&38.63&0.56&33.04\\  
         
         &&SMLM  & \textbf{84.96}& \textbf{78.05}&\textbf{67.40}&\textbf{53.61} & \textbf{0.57}&\textbf{20.66}\\  
    &&SAMLM & 84.75&77.98 &60.04&43.00&0.57&29.15 \\
  \hline
\end{tabular}}
\caption{Results of adversarial attack for setup 2. Here, PR: perturbation ratio (higher the better), PDA: percentage drop in accuracy (lower the better)} 
 \label{tab:attack_ind}
\end{table*}
\subsubsection{\textbf{Performance Evaluation on the Original Test Sets:}}
We evaluate the effectiveness of our proposed approaches on the original test sets of Hinglish and Benglish languages. Results of Hinglish are presented in Table \ref{tab:attack_ind} (BA and BF1). Our proposed pre-training approaches (SMLM and SAMLM) results in the improvement in classification tasks across the two code-mixed languages. In case of Hinglish sentiment and offensive classification tasks, the SAMLM pre-trained BERT model gives the best scores. 

Interestingly in Benglish dataset (c.f. Table \ref{tab:benglish_test}) aggression classification task our SMLM pre-training results in better classification. This may be because our Benglish code-mixed data consists of Bengali script words along with Romanized Bengali words. The SOUNDEX algorithm is unable to produce sound encodings for such words. Since SAMLM interleaves words and sound encodings, there are randomly missing sound encodings in a sequence that negatively affects alignment. SMLM- on the other hand- is not severely affected by the missing sound encoding as it does not explicitly align the word and sound encoding sequences.
We follow a paired T-test (significance test), which validates the performance gain over the baselines is significant with 95\% confidence (p-value$<$0.05). We observe that the gain over BA and BF1 is incremental whereas gain in AA and AF1 is larger. It is due to the fact that BA and BF1 are calculated on the original test sets and due to the small number of spelling variations in the original test set, gain is incremental. However, in the case of AA and AF1, there are more spelling variations in the adversarial test set. Larger gain in AA and AF1 illustrates the fact that our proposed approaches has the potential to handle these phonetic perturbations compared to other baselines.
More experiments are present in Appendix \ref{sec:appendix:moreexp}. 

\begin{table}
\small
   \resizebox{0.45\textwidth}{!}
{
  \begin{tabular}{cclll}
     \hline
    \textbf{Model}& \textbf{Task} &\textbf{Model}& \textbf{Accuracy} &\textbf{F1}\\
   \hline
    BERT &Aggression&VCBERT &  68.75&63.60 \\
     &&VMLMBERT  & 70.75&65.32 \\
       &&PhoneMLM  &70.65&66.01  \\ 
             
         &&SMLM &\textbf{75.89}&\textbf{67.52} \\  
     &&SAMLM &70.98&66.91\\

RoBERTa & Aggression & VCRoBERTa & 66.40&61.27 \\
     && VMLMRoBERTa  & 69.15&59.24 \\
       &&PhoneMLM &69.64 &61.63 \\  
       
         &&SMLM & \textbf{71.00} & 62.16 \\  
    &&SAMLM &68.08&\textbf{62.20}\\

  \hline
  
\end{tabular}
} \caption{Results on the original test set (non-adversarial) for Benglish}
 \label{tab:benglish_test} 
\end{table}

\section{Qualitative Analysis} 
This section analyzes the actual decision process of the proposed framework for the classification tasks by extracting the terms responsible for predicting the final output class. We explain the behaviour of different BERT-base models for Hinglish sentiment dataset. 

\subsection{Explaining Adversarial Robustness}
\begin{figure}[h]
  \centering
  \includegraphics[scale=0.37]{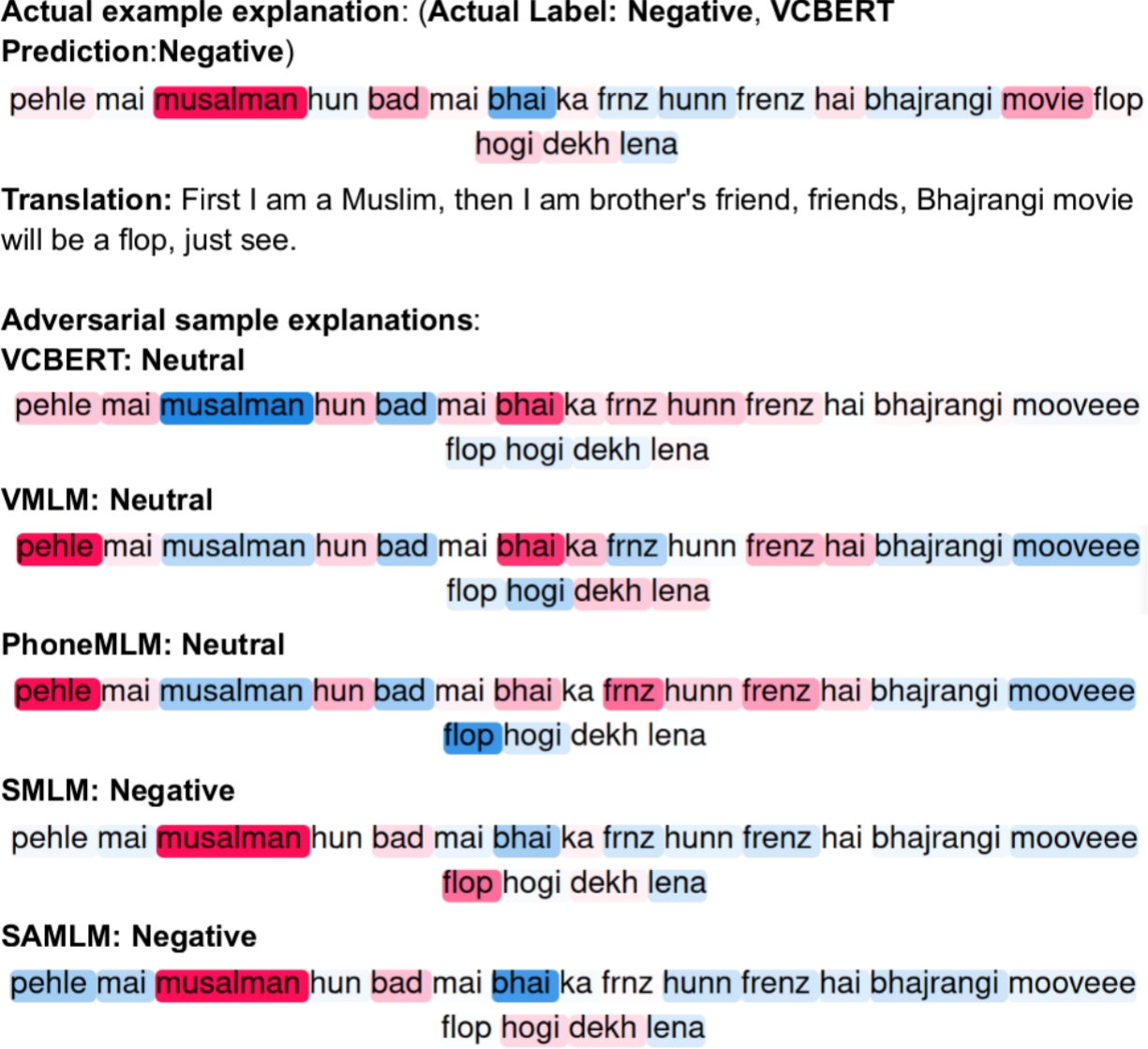}
  \caption{Qualitative Analysis for adversarial attack samples on the different BERT models}

 \label{ex1_crop}
\end{figure}

In this section, we explain how the auditory features help the model in improving its robustness \footnote{Robustness criteria defined in Section \ref{sec:threat}}. Figure \ref{ex1_crop} shows example of the Hinglish sentiment dataset where the predictions of all the models are affected due to the adversarial attack, but our model is robust for this attack. 
Tokens with red colour signify the terms which are responsible for the final label prediction (positive SHAP scores). In contrast, the words with blue colour negatively influence the final prediction (negative SHAP scores). More intense colour signifies the greater influence of the term for the final prediction.  
In Figure \ref{ex1_crop}, the actual label of the sentence is negative. Applying adversarial perturbation to the original example results in a successful attack against the VCBERT model, VMLM, and PhoneMLM models. However, SOUNDEX encoding helps SMLM and SAMLM to defend against this adversarial attack. Figure \ref{ex1_crop} reveals that in the case of the original example, words \text{musalman} (muslim), 
\textit{bad} (after), \textit{movie}, \textit{flop}, etc. are contributing positively for \textit{negative} sentiment prediction. However, words \textit{bhai} (brother), \textit{frnz} (friends), etc. contribute negatively to the \textit{negative} sentiment prediction. Adversarial attack on the VCBERT model by applying perturbations on word \textit{movie} (moovee) has shifted the focus from positively contributing words to other words. This behavior results in misclassification to \textit{neutral} class. On the other hand, SOUNDEX encoding helps the model to resist adversarial attack by assigning the same SOUNDEX encoding ($M100$) to \textit{movie} and \textit{moovee}. The same SOUNDEX encoding forces the model to treat both spelling variations equally. This shows that our proposed SMLM and SAMLM are more robust to such adversarial attacks. 

\subsection{Explaining Text Classification}
\begin{figure}[t]
  \centering
  \includegraphics[scale=0.32]{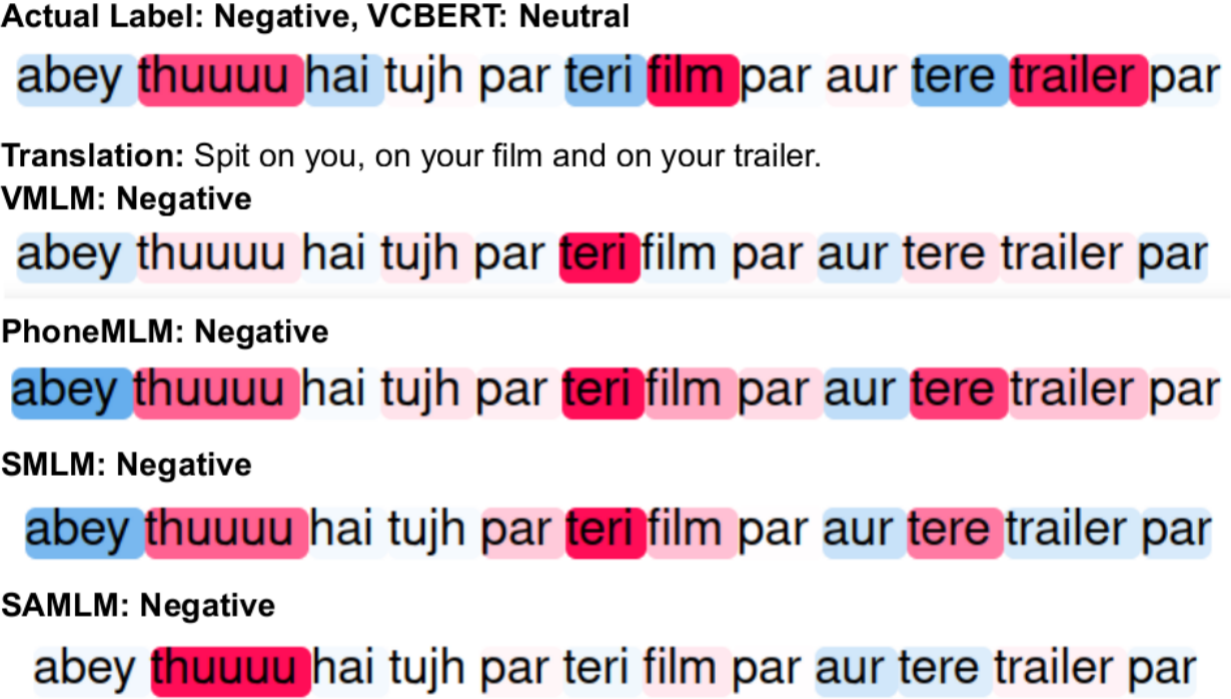}
  \caption{Qualitative analysis for Hinglish text classification using BERT based models}
  \label{ex1_clasi}
\end{figure}

\begin{figure}[h]
  \centering
  \includegraphics[scale=0.29]{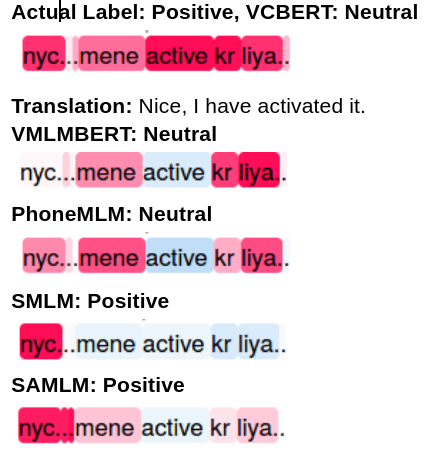}
  \caption{Qualitative Analysis for Hinglish text classification using BERT based mdoels.}
 \label{clasi_pos}
\end{figure}

We discuss how the addition of auditory features and our pre-training mechanisms helps the classifiers in improving their performance. 
We show a few examples where (i). the VCBERT model does misclassification, but all the other models produce the correct classes (Figure \ref{ex1_clasi}); (ii). VCBERT, PhoneMLM BERT, and VMLMBERT perform misclassification but our proposed models perform correct classification (Figure \ref{clasi_pos}).

In Figure \ref{ex1_clasi}, although the VCBERT focuses on the word \textit{thuuuuu} (spit) but due to the repetition of character \textit{u}, it is not able to understand it. VCBERT commits 
misclassification due to focus on the other words \textit{film} and \textit{trailer}. On the other hand, all the other models are able to capture these spelling variations and perform correct classification.

Figure \ref{clasi_pos} illustrates the case where all other models misclassify but our proposed SMLM and SAMLM correctly classify. Although the focus of VCBERT, VMLMBERT, and PhoneMLM BERT models are on the \textit{nyc} (spelling variation of \textit{nice}), but these models are not able to identify it, and as a result it turns into a 
mislcassification to neutral class. Our proposed approaches, SMLM and SAMLM perform correct classification by assigning the same encoding to \textit{nice} and \textit{nyc} (N200). This illustrates the effectiveness of SOUNDEX encoding and our proposed SMLM and SAMLM pre-training in capturing the spelling variations more effectively than the baselines. More detailed analysis is present in Appendix \ref{sec:appendix:shap}.

\begin{table*}[hbt]
\small

   \resizebox{0.9\textwidth}{!}
{
  \begin{tabular}{lp{0.37\linewidth}lllll}
    \hline
   \textbf{ Sr No.} & \textbf{Example} & \textbf{Actual Label} & \textbf{SMLM} & \textbf{SAMLM} \\
    \hline
    1 & W8 is ovr & neutral &positive & positive \\
    & \textit{Translation:} Wait is over.\\ 
  2&  Sir apke liye puri life wait kar sakte h &positive &neutral &neutral \\
  & \textit{Translation: Sir can wait whole life for you} &\\
  3& Bhai apka being humen trust kesa chal rha he &neutral &positive &positive \\
  & \textit{Translation: Brother how is your being humen trust going} & \\
  \hline
\end{tabular}}
 \caption{Error analysis for Hinglish text classification (original test set) using BERT-based models}
  \label{tab:textclasierror}
\end{table*}

\subsection{Error Analysis}
\label{apx:sec:error}
To explain the limitations of our proposed framework, we show samples misclassified by the SMLM and SAMLM models in Table \ref{tab:textclasierror}. Samples are taken from Hinglish language sentiment classification task (BERT based models).
In example 1, the word \textit{wait} is written as \textit{W8}. Here SOUNDEX algorithm encodes it as W000 (numbers are not captured by SOUNDEX algorithm). Hence, both models randomly predict the positive sentiment.
Example 2 has implicit positive sentiment which both the SMLM and SAMLM models are unable to understand, resulting in misclassifications. The VCBERT, VMLMBERT, and PhoneMLMBERT models also misclassify such samples. In example 3, SMLM and SAMLM models predict positive sentiment because both models focus on the words bhai (brother) and trust (revealed by SHAP). The presence of these words has created confusion for both models, which is the reason for misclassification.

\section{Conclusion}
\label{sec:conclude}
In this paper, we propose two novel pre-training steps, SMLM (SOUNDEX Masked Language Modeling) and SAMLM (SOUNDEX Aligned Masked Language Modeling), to incorporate the auditory phonetic (AP) features into popular classification models, BERT and RoBERTa. Our approach effectively handles spelling perturbations, a common form of attack in code-mixed languages like Hinglish and Benglish. We perform phonetic-based adversarial attacks on models trained using our technique and find that the performance decrease is significantly less than multiple baselines. Additionally, incorporating the AP features leads to improvement in classification scores on different tasks in both Hinglish and Benglish as compared to models trained only on semantic features. In summary, the novel pre-training steps of SMLM and SAMLM provide an effective way to incorporate AP features into NLP models, leading to improved robustness and performance on code-mixed text classification tasks.

In future work, we plan to extend our approach to other code-mixed languages and evaluate its performance on more NLP tasks. We believe that our approach can have a significant impact on the robustness of NLP models, especially in the context of code-mixed languages.

\section*{Limitations}
This study, like most studies, has some limitations that could be addressed in future research. Our approach does not fix the issue of implicit sentiment in sentences that are present in the corresponding baseline models. SOUNDEX does not give encoding for numeric digits resulting in the same representation for different words containing such digits. For such words, our approach would not give any boost in performance over the baselines. We have discussed such examples in Section \ref{apx:sec:error}. In addition, our proposed approach can not handle code-mixed languages written in original script. 
These limitations could be addressed in the future by augmenting more data of an implicit nature through the semi-supervised way and through the better encoding of auditory features.

\section*{Ethics Statement}
We use freely available datasets for our experiments. The
dataset has been used only for academic purposes,
and in complete compliance with the license.

\section*{Acknowledgement}
Authors would like greatly acknowledge ''Centre for Development of Telematics, India (C-DOT)'' for its partial support to carry out this research.

\bibliography{anthology,custom}
\bibliographystyle{acl_natbib}

\appendix

\section{Implementation and Dataset Details}
\label{sec:appendix:impdata}
\subsection{Implementation Details}
We use BERT-base and RoBERTa-base as target models for each task. To implement our models, we use the Python-based library Pytorch \footnote{https://pytorch.org/} and Hugging face implementation of BERT and RoBERTa \cite{wolf2019huggingface}.  
Target model BERT-base uses 12 layers of transformers block with a hidden size of 768 and number of self-attention heads as 12. It has 110M trainable parameters. 
RoBERTa-base is pre-trained on a large corpus of English data in a self-supervised fashion. It has a hidden size of 768 and contains 12 hidden layers. RoBERTa-base model has 125M trainable parameters.
We use the Adam optimizer to optimize the network and the weights update is computed based on the categorical cross-entropy loss for all the classification tasks. 
The hyper-parameters of both models are also fine-tuned for both languages on the respective task datasets. We use the grid search to find the best set of hyper-parameters. We perform all the computations on the Nvidia929GeForce GTX 1080 GPU with 12 GB memory.

\subsubsection{Computational Efficiency}
\label{apx:sec:compu}
Our proposed approach is computationally less expensive than the existing adversarial approaches. 
The existing adversarial approaches use adversarial training to increase the robustness of the system, which involves computationally expensive steps of generating the adversarial samples from the training set. Adversarial examples are generated by first finding the importance of every word in the sentence and then applying perturbations to important words until the attack is successful. Suppose there are $n$ words in the sentence; then this traditional approach requires $n$ number of queries to the trained model to calculate the importance of each word. Further, more queries are required to generate adversarial samples. In the worst case, $n$ number of operations are required on actual example (perturbing each word of example to execute a successful attack). This will again require $n$ queries to the trained model. This process is computationally expensive and requires $MxNxN$ computations ($M$ number of instances in the training step, $N$ average token length of each instance). The existing model is further fine-tuned on these adversarial samples to make it robust. Our approach gets rid of this computationally expensive process by introducing a small pre-training step as discussed in Section \ref{sec:method}. Both our approaches, SMLM and SAMLM do not require pre-training a model from scratch, but only require a small pre-training step (before final fine-tuning) utilizing a very few instances (33,014 Hinglish and 6,149 Benglish in our case) on the existing pre-trained language models. 
As a result of SMLM and SAMLM, our approach does not require re-training (adversarial training) of the classifier on adversarial test samples.

\subsection{Datasets}
To access the experimental evaluation of our proposed approach, we conduct extensive experiments on code-mixed Hinglish and Benglish language datasets. For Hinglish, we conduct experiments on two benchmark datasets related to offensive and sentiment analysis. For Benglish, we conduct experiments on aggression analysis data. Details of the datasets are described below:
\paragraph{\textbf{Hinglish Sentiment Analysis Dataset \cite{joshi2016towards}:}} 
This dataset contains posts from some public Facebook pages popular in India. The dataset is annotated with three sentiment classes, \textit{viz.}, positive, negative, and neutral. It contains a total of 3,879 instances.
\paragraph{\textbf{Hinglish Offensive Tweet (HOT) Dataset \cite{mathur2018did}:} }
HOT dataset contains tweets crawled using Twitter Streaming API by selecting tweets having more than three Hinglish words. It is manually annotated with 3 classes, \textit{viz.}, non-offensive, abusive, and hate-inducing. This dataset contains a total of 3189 tweets.

\paragraph{\textbf{Benglish Aggression Analysis Dataset \cite{bhattacharya2020developing}:}}
This dataset is collected from comments on YouTube comments and contains comments written in Bengali as well as Roman scripts. It contains 5971 comments, annotated with 3 classes of aggression, \textit{viz.}, overtly aggressive, covertly
aggressive, and non-aggressive.

All the datasets are divided into 3 splits- train, validation, and test. 
The detailed statistics of all the dataset splits are shown in Table \ref{tab:stat}.

\paragraph{\textbf{Pre-training Datasets:}}
\begin{itemize}[noitemsep]
    \item \textit{Hinglish:} We pre-train the models on a total of 33,014 Hinglish sentences. We utilize the publicly available code-mixed datasets from \citet{joshi2016towards, mathur2018did, patwa2020sentimix} for this task. In addition, we also crawled 9,141 tweets from Twitter using Search API \footnote{https://developer.twitter.com/en/docs/twitter-api/v1/tweets/search/api-reference/get-search-tweets} and added them to our pre-training corpus.
    
    \item \textit{Benglish:} SMLM and SAMLM are pre-trained on 6,149 code-mixed sentences taken from the publicly available datasets by \citet{bhattacharya2020developing} and \citet{jamatia2015part}.
\end{itemize}
These pre-training datasets are divided into 2 splits- train (80\%) and validation (20\%).

\begin{table*}
\centering
  \begin{tabular}{cclll}
    \hline
    \textbf{Language}& \textbf{Task} &\textbf{Train}& \textbf{Development} &\textbf{Test}\\
   \hline
    Hinglish &Sentiment&2482  & 621&776\\
     &Offensive&2710&271&478 \\
     Benglish  &Aggression& 5075&508&896  \\ 
 \hline
\end{tabular}

\caption{Data Statistics}
 \label{tab:stat}
\end{table*}

\section{More Experiments}
\label{sec:appendix:moreexp}
To demonstrate the effectiveness of passing SOUNDEX vectors along with textual content, we perform experiments for setup 2 (defined in Section \ref{sec:exp}), which involves performance evaluation on the original test sets. Since the vanilla pre-trained models of BERT and RoBERTa do not incorporate any SOUNDEX information, fine-tuning these models with only SOUNDEX vectors would be unfair. Therefore, we experiment on SMLM and SAMLM that have a SOUNDEX component in pre-training. We pass only the SOUNDEX vectors to both the models during task fine-tuning for Hinglish and Benglish languages. Evaluation results for Hinglish and Benglish language tasks are shown in Table \ref{tab:resu}.
\begin{table*}
   \resizebox{0.99\textwidth}{!}
{
\centering
  \begin{tabular}{cclllll}
    \hline
    \textbf{Language}& \textbf{Task} &\textbf{Model}& \textbf{Accuracy (SOUNDEX)} &\textbf{F1 (SOUNDEX)} & \textbf{Accuracy (SOUNDEX $+$ Text)} &\textbf{F1 (SOUNDEX $+$ Text)}   \\
   \hline
    Hinglish &Sentiment&SMLM   & 61.47&52.21&\textbf{ 69.33} &\textbf{65.30}\\
    & &SAMLM &63.46 &54.17 &\textbf{70.36} &\textbf{67.56} \\
     &Offensive& SMLM&81.38&74.53 &\textbf{88.31}&\textbf{82.20} \\
     & & SAMLM & 82.00&74.40&\textbf{88.93}&\textbf{83.31} \\
     Benglish  &Aggression& SMLM  &61.99&55.51 &\textbf{75.89}&\textbf{67.52} \\ 
     & & SAMLM & 62.43&56.19&\textbf{70.98} &\textbf{66.91}\\
 \hline
\end{tabular}}
\caption{Results on the original (non-adversarial) test set for Hinglish and Benglish for BERT-based models}
 \label{tab:resu}
\end{table*}

We observe that using only SOUNDEX vectors performs inferior compared to our proposed approach, where we are passing SOUNDEX vector along with semantic features. In this case, SOUNDEX will assign the same encoding vectors (Y600) to the Hindi word \textit{yar} (friend) and English word \textit{year}. These cases will add to the model's confusion, which could be the possible reason for its inferior performance. In our proposed approach, this limitation of SOUNDEX is handled by providing the word tokens along with SOUNDEX tokens at the input.

\subsection{Evaluating Multilingual Models}
We also perform experiments to assess the robustness of multilingual models. We perform experiments with multilingual BERT (mBERT) and IndicBERT for sentiment classification for the Hinglish language. The mBERT and InidcBERT obtain accuracy of 65.31\% and 49\%, respectively, for the sentiment Hinglish task. We further perform detailed experiments to assess the robustness of mBERT and IndicBERT against adversarial attacks. Adversarial attack is performed using subword perturbations as described in Section \ref{sec:evaladv}.
\subsubsection{Evaluation Results on Adversarial Test Sets}

We define two setups (similar to Section \ref{sec:evaladv}): (i). generate attack samples by attacking vanilla mBERT and VCIndicBERT (vanilla IndicBERT) and evaluate the performance on all other models; (ii). attack individual models by generating different adversarial samples for each model.
Results for setup 1 and setup 2 are depicted in Tables \ref{tab:mt_bert_setup1} and \ref{tab:mt_bert_setup2}, respectively.

Similar phenomena have been observed in the case of mBERT and IndicBERT multilingual models, mirroring the observations made for BERT-base and RoBERTa-base models. We observe that mBERT and InidcBERT models are also vulnerable to phonetic perturbations based adversarial attack. It is evident from the larger drop in accuracy and F1 scores for setup 1 and setup 2. Our proposed pre-training approaches, SMLM and SAMLM illustrate their effectiveness in improving the robustness of mBERT and IndicBERT models by minimizing the drop in accuracy and F1 scores compared to other baselines.

\subsubsection{Evaluation Results on Original Test Sets}
We test the effectiveness of our proposed approach on original test sets of Hinglish sentiment task. Results are presented in Table \ref{tab:mt_bert_setup2} (BA and BF1). It is observed that our pre-training approaches help the mBERT and IndicBERT models in improving their performance. Our proposed approaches give better results for both models, illustrating the importance of using auditory features. 
It is also observed that although IndicBERT models (all variations) achieve high accuracy, the F1 score is very low compared to the mBERT model. It is due to low class-wise performance.  

\begin{table}[t]
\small
 \resizebox{0.49\textwidth}{!}
{
  \begin{tabular}{cclllllll}
  
   \hline
    \textbf{Attack Model}& \textbf{Task} &\textbf{Model}& \textbf{AA} & \textbf{AF1} 
    \\
   \hline
    mBERT &Sentiment&VCBERT  &42.78&26.33 \\
     Acc=65.31, F1=61.16&&VMLMBERT & 60.97&57.88\\
       &&PhoneMLM & 60.49&57.59\\             
         &&SMLM   & 64.30&59.53\\
    &&SAMLM & \textbf{64.69} & \textbf{59.85}\\
    
IndicBERT &Sentiment&VCIndicBERT & 35.18&22.80 \\
     55.02, 43.85&&VMLMIndicBERT & 49.04&34.54\\
       &&PhoneMLM & 49.69&35.77\\  
      
         &&SMLM  & 51.08&40.44\\
    &&SAMLM &\textbf{51.18}&\textbf{40.68}\\
 \hline
\end{tabular}}
 \caption{Results for adversarial attack (samples generated by attacking original multilingual BERT/IndicBERT models)}
  \label{tab:mt_bert_setup1}
\end{table}

\begin{table*}[t]
\small
\centering
  \begin{tabular}{ccllllllll}
    \hline
    \textbf{Model}& \textbf{Task} &\textbf{Model}& \textbf{BA} &\textbf{BF1} & \textbf{AA} &\textbf{AF1} &\textbf{PDA}
    \\
    \hline
    mBERT &Sentiment&VCBERT & 65.31&61.16&42.78&26.33&52.66 \\ 
     &&VMLMBERT  &66.23&61.03&44.20&38.43&33.26\\ 
       &&PhoneMLM &  66.55&61.32 & 42.66&35.43 & 35.89\\ 
            
         &&SMLM  &67.26&63.44 &53.35&44.64&20.68  \\ 
    &&SAMLM &\textbf{67.91} &\textbf{65.38}& 58.89&52.77&\textbf{13.28}\\
    
 IndicBERT &Offensive&VCBERT& 55.02 & 43.85 &35.18&22.81 & 36.05\\
     &&VMLMBERT  & 63.78&45.10 &45.87&28.27 & 28.08\\
       &&PhoneMLM &64.01&46.63&46.11&29.36&27.96 \\  
      
         &&SMLM  &\textbf{ 65.59}&\textbf{52.33} & 52.03&35.40 &20.67\\
    &&SAMLM & 63.65&49.10 & 52.96&33.43 &\textbf{ 16.79}\\
    
  \hline
\end{tabular}
\caption{Results of adversarial attack (generating adversarial samples against individual model). Here, PR: perturbation ratio (higher the better), PDA: percentage drop in accuracy (lower the better)} 
 \label{tab:mt_bert_setup2}
\end{table*}

\subsection{Language Generalizability}
Our approach can be generalized to other code-mixed languages written in Romanized script. In general, this approach can be applied to any language where the Romanization of native script leads to spelling variations. Hindi and Bengali languages, when written in Romanized code-mixed form, produce many such spelling variations. Similarly, Punjabi belongs to the same language family, and the Romanized code-mixing form of Punjabi also induces spelling variations in the data. We perform additional experiments with Punjabi-English code-mixed language to demonstrate the generalizability capability of our proposed approach.
 We use the publicly available dataset for sentiment task to evaluate our model on robustness and accuracy metrics (explained in Section \ref{sec:threat}) \cite{yadav2020bilingual}. Experimental results for Punjabi-English corresponding to setup 2 are presented in Table \ref{tab:mt_bert_setup2_punj} (setup2). We observe that auditory features help Punjabi-English language to improve robustness and accuracy, similar to other language pairs. 

\begin{table*}[t]
\small
\centering
  \begin{tabular}{ccllllllll}
    \hline
    \textbf{Model}& \textbf{Task} &\textbf{Model}& \textbf{BA} &\textbf{BF1} & \textbf{AA} &\textbf{AF1} &\textbf{PDA}
    \\
    \hline
      BERT &Punjabi-&VCBERT & 71.14 &68.09 &26.40&25.20 & 62.89 \\
     &English&VMLMBERT  &72.66& 69.71 &39.81&37.31& 45.21\\
       &&PhoneMLM & 72.31&69.88& 38.03&37.65&45.40\\ 
            
         &&SMLM  & 73.21&70.41&42.35 & 40.32& 42.15\\  
    &&SAMLM & \textbf{73.59}&\textbf{70.51}& \textbf{54.88} &\textbf{53.18}&  \textbf{25.42}\\

  \hline
\end{tabular}
\caption{Results for adversarial attack for Punjabi-English Pair (generating adversarial samples against individual model). Here, PDA: percentage drop in accuracy (lower the better)} 
 \label{tab:mt_bert_setup2_punj}
\end{table*}

\section{Qualitative Analysis}
\label{sec:appendix:shap}
\subsection{Explaining Adversarial Robustness}
\begin{figure}[t]
  \centering
  \includegraphics[scale=0.37]{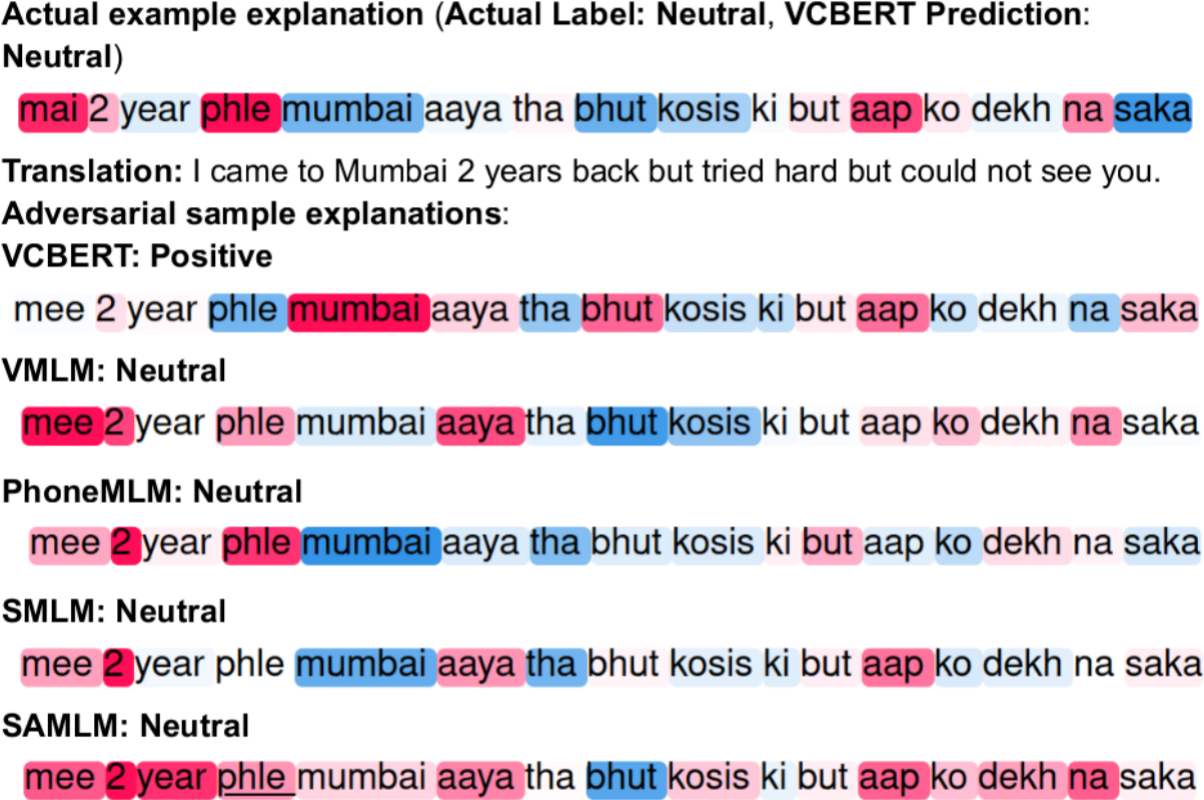}
  \caption{Qualitative Analysis for adversarial attack samples on the different BERT models.}
   \label{ex2_crop}
\end{figure}

In this section, we explain how the auditory features help the model in improving its robustness. Figure \ref{ex2_crop} show examples of the Hinglish sentiment dataset where the predictions of VC models are affected due to the adversarial attack. 
In example 1 (\ref{ex2_crop}), replacing \textit{mai} (I) with \textit{mee} causes the vanilla BERT model to perform misclassification. However, all other models are robust. Figure \ref{ex2_crop} explains the decision process of all the models. Tokens with red colour signify the terms which are responsible for the final label prediction (positive SHAP scores). In contrast, the words with blue colour negatively influence the final prediction (negative SHAP scores). More intense colour signifies the greater influence of the term for the final prediction.

Figure \ref{ex2_crop} reveals that for predicting the neutral sentiment for actual example 1, original BERT focuses more on words \textit{mai} (I) and \textit{phle} (before) and words \textit{Mumbai}, \textit{bhut} (very) and \textit{saka} (did) makes a negative impact for the neutral class classification. Changing the word \textit{mai} to \textit{mee} (adversarial example) shifts the focus of original BERT to other words like \textit{Mumbai}, \textit{bhut} (very), \textit{aap} (you), etc. This shift of focus to negatively contributing words results in increasing confusion for the BERT model which is the reason for misclassification. However, MLM, PhoneMLM, SMLM and SAMLM help the BERT model to keep its focus on positively contributing words. Here, SMLM and SAMLM will assign same the encoding vector to \textit{mai} and \textit{mee} $(M000)$ which help the model to defend against adversarial attack. 

\begin{figure}[t]
  \centering
  \includegraphics[scale=0.34]{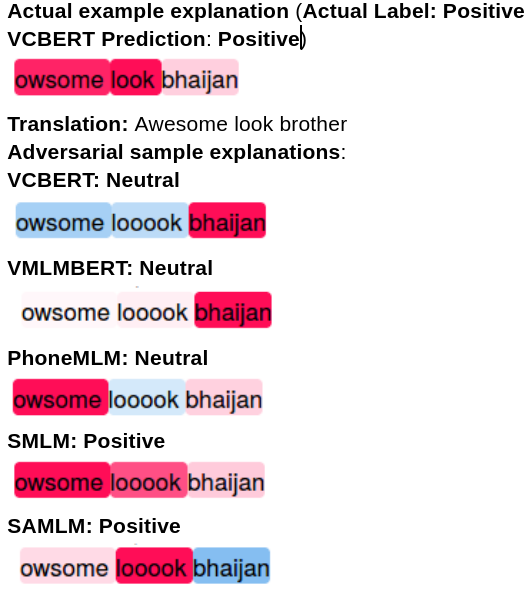}
  \caption{Qualitative Analysis for adversarial attack samples on the different BERT models.}
   \label{ex_pos}
\end{figure}

Figure \ref{ex_pos} shows example of case where an adversary is able to execute a successful attack against all the models except SMLM and SAMLM.
Here, the main focus of VCBERT model is on \textit{owesome} (variation of awesone) and look. After changing the word \textit{look} to \textit{looook}, the focus of VCBERT and VMLMBERT have shifted to \textit{bhaijan} (brother), which results in misclassification to neutral class. In case of PhoneMLM, model's focus is on \textit{owsome} and \textit{bhaijan} (light red). However, the word \textit{looook} now negatively contributes as the model is not able to recognize it, and this in turn results in misclassification. Our proposed approaches, SMLM and SAMLM are able to recognize this spelling variation, and hence classify correctly.

\subsection{Explaining Text Classification}
\begin{figure}[t]
  \centering
  \includegraphics[scale=0.37]{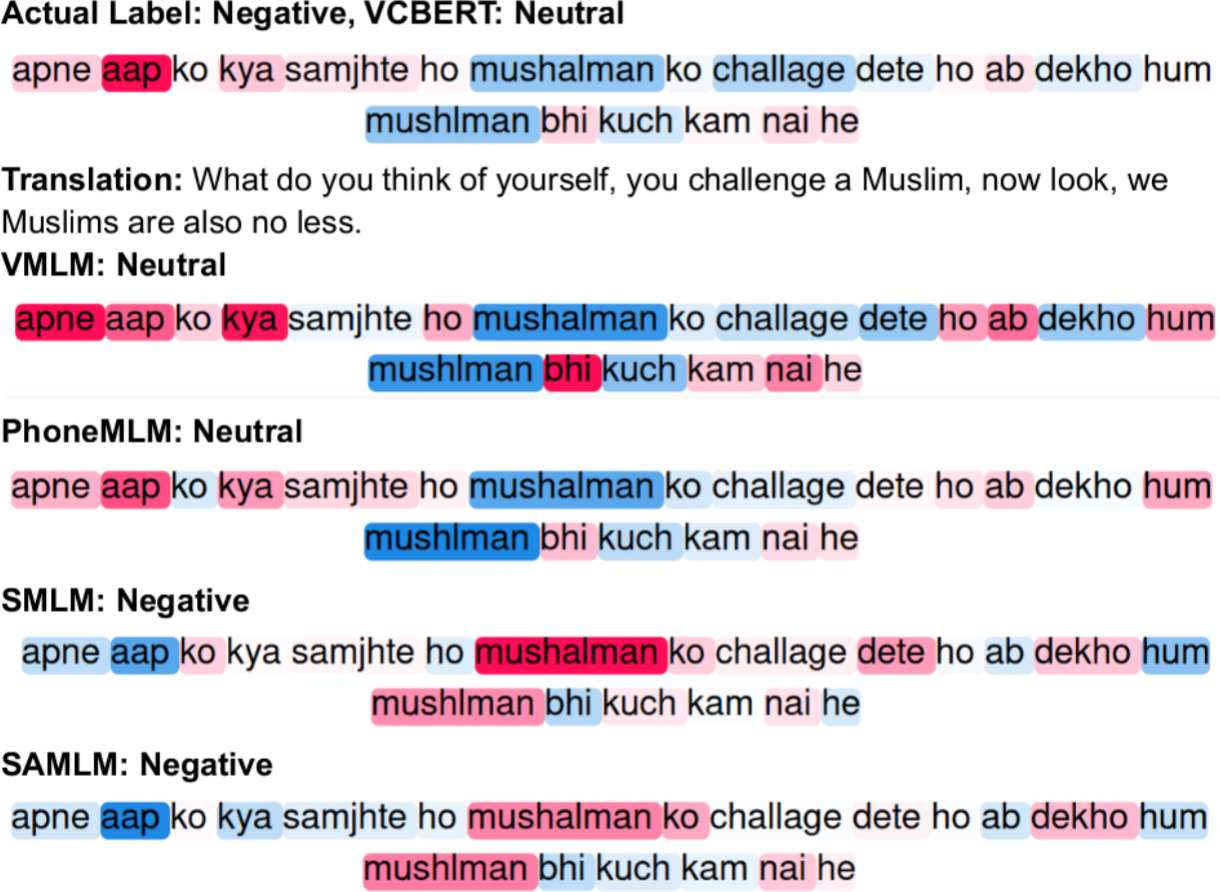}
  \caption{Qualitative analysis for Hinglish text classification using BERT based models.}
   \label{ex3_clasi}
\end{figure}

Figure \ref{ex3_clasi} illustrates the case where all other models perform misclassification but our proposed SMLM and SAMLM perform correct classification. The VCBERT, VMLMBERT, and PhoneMLM BERT perform misclassification due to their wrong focus on neutral words \textit{aap} (you), \textit{kya} (what), \textit{ab} (now), \textit{hum} (we), etc. These models, including PhoneMLM are not able to focus on the correct words due to spelling variations in them. However, our proposed model is able to capture these variations in words \textit{challage} (spelling variation of word \textit{challenge}) and context word mushalan (muslim) (spelling variation of word \textit{musalman}), nai (spelling variation of word \textit{nahi}), etc. It will assign same sound encoding $C420$ to both \textit{challenge} and \textit{challage}, $M245$ to \textit{mushalan} and \textit{musalman}, $N000$ to \textit{nai} and \textit{nahi}. 

\end{document}